\documentclass[pmlr]{jmlr}% new name PMLR (Proceedings of Machine Learning)

\makeatletter
\def\set@curr@file#1{\def\@curr@file{#1}} %temp workaround for 2019 latex release
\makeatother
\usepackage[load-configurations=version-1]{siunitx} % newer version
\usepackage{lipsum}
\usepackage{mwe}
\usepackage{tikz}
\makeatletter
\newcommand*{\centernot}{%
  \mathpalette\@centernot
}
\def\@centernot#1#2{%
  \mathrel{%
    \rlap{%
      \settowidth\dimen@{$\m@th#1{#2}$}%
      \kern.5\dimen@
      \settowidth\dimen@{$\m@th#1=$}%
      \kern-.5\dimen@
      $\m@th#1\not$%
    }%
    {#2}%
  }%
}
\makeatother

\usepackage{mathtools}

\DeclarePairedDelimiterX\Basics[1](){ #1}

\usepackage{algorithm}
\usepackage{algorithmic}
\usepackage{enumitem}
\usepackage{multirow}

\usepackage{longtable}% for long tables

\usepackage{booktabs}
\usepackage[load-configurations=version-1]{siunitx} % newer version
\theorembodyfont{\upshape}
\theoremheaderfont{\scshape}
\theorempostheader{:}
\theoremsep{\newline}

\jmlrvolume{XXX}
\jmlryear{2022}
\jmlrworkshop{Machine Learning for Healthcare}

\title[Causal Inference on Delirium Patients from MIMIC-III]{Causal Discovery on the Effect of Antipsychotic Drugs on Delirium Patients in the ICU using Large EHR Dataset}

\author{\Name{Riddhiman Adib} \Email{riddhiman.adib@marquette.edu} \\
      \addr Department of Computer Science\\
      Marquette University\\
      Milwaukee, Wisconsin, USA
      \AND
      \Name{Md Osman Gani} \Email{mogani@umbc.edu} \\
      \addr Department of Information Systems\\
      University of Maryland, Baltimore County\\
      Baltimore, Maryland, USA
      \AND
      \Name{Sheikh Iqbal Ahamed} \Email{sheikh.ahamed@marquette.edu} \\
      \addr Department of Computer Science\\
      Marquette University\\
      Milwaukee, Wisconsin, USA
      \AND
      \Name{Mohammad Adibuzzaman} \Email{adibuzza@ohsu.edu} \\
      \addr Oregon Clinical and Translational Research Institute\\
      Oregon Health \& Science University\\
      Portland, Oregon, USA
}

\begin{document}

\maketitle

% Tells us a bit about the problem.  Recent advances in machine learning
% \citep{cite1} have resulted in great things happening in healthcare.
% In particular, \citet{cite2} describes a spiffy technique to save even
% more lives.  In this work, we...

% ARXIV VERSION
\begin{abstract}
    Delirium occurs in about 80\% cases in the Intensive Care Unit (ICU) and is associated with a longer hospital stay, increased mortality and other related issues. Delirium does not have any biomarker-based diagnosis and is commonly treated with antipsychotic drugs (APD). However, multiple studies have shown controversy over the efficacy or safety of APD in treating delirium. Since randomized controlled trials (RCT) are costly and time-expensive, we aim to approach the research question of the efficacy of APD in the treatment of delirium using retrospective cohort analysis. We plan to use the Causal inference framework to look for the underlying causal structure model, leveraging the availability of large observational data on ICU patients. To explore safety outcomes associated with APD, we aim to build a causal model for delirium in the ICU using large observational data sets connecting various covariates correlated with delirium. We utilized the MIMIC III database, an extensive electronic health records (EHR) dataset with 53,423 distinct hospital admissions. Our null hypothesis is: there is no significant difference in outcomes for delirium patients under different drug-group in the ICU. Through our exploratory, machine learning based and causal analysis, we had findings such as: mean length-of-stay and max length-of-stay is higher for patients in Haloperidol drug group, and haloperidol group has a higher rate of death in a year compared to other two-groups. Our generated causal model explicitly shows the functional relationships between different covariates. For future work, we plan to do time-varying analysis on the dataset. 
\end{abstract}

\section{Background and Problem Statement}

% Delirium, importance, lack of agreement in studies

% Antipsychotic drugs \textit{(Haloperidol, Ziprasidone, Olanzapine, etc.)}
% Estimating the effect of experimentation (doing) is not feasible based only on observational data (seeing)

With a focus on the theoretical development of causal inference methodologies in the previous three chapters, this chapter aims to discuss a practical, real-world application of the causal inference framework to untangle unknown healthcare information. For this purpose, we look into Delirium patients in the ICU. 

Delirium (or acute brain failure) \cite{girard2008delirium} is a disorder or disruption of consciousness, presenting with a reduced capacity to focus, sustain, or shift concentration. Delirium occurs in about 80\% cases in the Intensive Care Unit (ICU) and is associated with a more extended hospital stay, increased mortality for each additional day with Delirium in the ICU \cite{pisani2009days} and other clinical complications such as self-extubation and removal of catheters. Two of the significant issues in diagnosing and treating delirium patients are: 

\begin{itemize}
    \item Currently, no biomarker exists to diagnose Delirium; rather, Delirium is diagnosed with subjective assessment tools such as the confusion assessment method (CAM) \cite{inouye1990clarifying,wassenaar2015multinational}. This diagnosis requires the presence of a physician active in the medical center and makes the diagnosis and detection of Delirium patients in the real-world challenging.
    \item Delirium is commonly treated with antipsychotic drugs (APD) \cite{girard2008delirium} such as Haloperidol, Ziprasidone, Olanzapine, etc. However, multiple randomized controlled trials (RCTs) have shown either conflicting or inconclusive results about the efficacy of APD in the treatment of delirium \cite{neufeld2016antipsychotic,hatta2014antipsychotics}. This has created a controversy over the efficacy or safety of APD in treating Delirium. 
\end{itemize}

% * What is RCT [Looking for answers: RCT]
RCTs have been considered the gold standard since the 1960s \cite{greene2012reform}. The goal was to identify the causation of diseases and understand the causal effect of drugs by the regulatory bodies such as the FDA and clinical communities. The key ideas behind RCT are:
\begin{itemize}
    \item By random assignment of treatment or interventions, the confounding bias, i.e., the bias due to the assignment of treatment or presence of other variables, can be removed from the estimand, including the unobserved confounders.
    \item By comparing similar population groups of treatment and control arm, an estimation can be made about treatment efficacy in the target population group.
\end{itemize}

However, RCTs have their own set of challenges as well. RCTs have become increasingly time-consuming, costly, and are often infeasible for safety and efficacy reasons \cite{frieden2017evidence}. Thus there is a need to find alternatives to RCTs, possibly to detect causal effects from other sources of information and aid in removing controversies of treatments in the field.

% Why CI for RCT and experimental studies
% Large EHR Dataset: Observational Study \textit{(or simulated RCT)}
% Exploration of \textbf{observational dataset}
% Utilize previously described theories

\begin{figure}[htbp]
    \centering
    \includegraphics[width=0.3\textwidth]{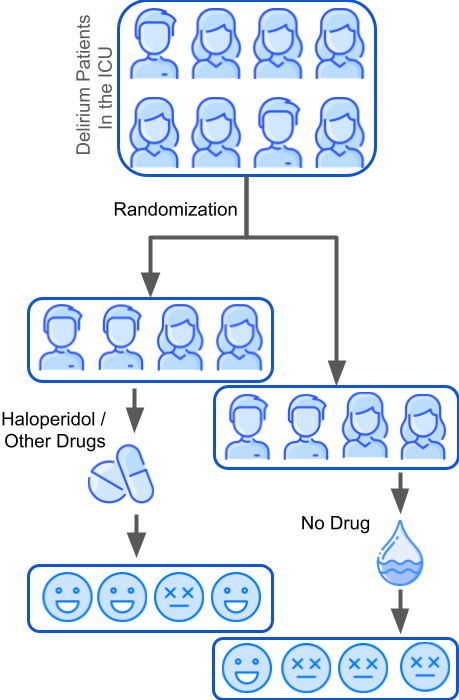}
    \caption{RCT for Antipsychotics-based treatment for Delirium}
    %  ~\label{fig:my_label}
\end{figure}

Recent advances in technology and the adoption of computerized systems in routine healthcare have enabled the collection and curation of large volumes of data during routine healthcare, albeit with confounding biases. At the same time, recent advances in the theory of causal model, more specifically structure causal models (SCM), provides the framework for adjusting for these confounding biases in many cases. This removal of biases can be done (sometimes even if the confounders are unobserved) from observational data using adjustment formulas such as backdoor/front door criterion \cite{bareinboim2016causal,pearl2016causal,spirtes2000causation}. However, this approach requires developing a graphical representation of the problem domain with meticulous scrutiny of the variables' relationship, structure learning algorithms, clinical experience, and existing literature. 

With the availability (collection, storage, maintenance) of large-scale data in different branches of science (healthcare, finance, sociology) and technology (connected health, smart home), opportunities exist to extract necessary information from it. Big data has aided in the revolution of neural networks, advanced reinforcement learning, and improved statistical machine learning methodologies. Most research advancements integrating big data revolve around curve fitting and correlation. However, without causal relationships, scientists lack the power of intervention or to even explore hypothetical scenarios (counterfactuals). 

Big data is responsible for many breakthroughs and advancements in healthcare, contributing to improved treatment policy solutions and collaborated information from multiple sources. Although most breakthroughs are based on predictive models, causal relationships are more crucial for healthcare. This has led to countless experimental trials (randomized controlled trials, case-control studies) on finding the efficacy or impact of an intervention on target outcomes. Causal inference leverages big data and contributes to finding causal information, sometimes even without experiments. One of the strengths of causal inference methodologies is to draw conclusions on causal effects from observational data. Causal inference and its potential with big data are not limited to healthcare only; it has shown great potential in other fields (finance, sociology, law) as well \cite{hofer2008injury,nishioku2009detachment,semmler2005systemic}. Artificial intelligence is iteratively improved with research work and is getting better at decision making and predictive modeling.

Since RCTs cost a lot in terms of money and time, emulation of RCTs from the observational dataset can help reduce them. It also would aid in using datasets from all over the world to find causation in other diseases and health complexities. While RCTs are the gold standard for identifying causal effects of interventions, it is time-consuming and costly. On the other hand, the data collected during routine care, such as electronic health records (EHR), might also be valuable to generate insight, identify the disease pathway and estimate the effect of interventions using recent advances in methods for causal inference.

% Our proposition: extraction of data from MIMIC and exploration (plus the CKH pipeline)
% Can we explore existing large observational healthcare data to find the efficacy (causal effect) of antipsychotics in the treatment of Delirium?
We aim to study the efficacy of APD in the treatment of Delirium using retrospective cohort analysis. We plan to use the Causal inference framework to look for the underlying causal structure model, leveraging the availability of large observational data on ICU patients. It will help us to untangle the causal relationship between variables and look into the counterfactual world (what-if). To explore safety outcomes associated with APD, our research work targets to develop a causal model for Delirium in the ICU using large observational data sets. We aim to utilize the MIMIC III database, an extensive electronic health records (EHR) dataset with 53,423 distinct hospital admissions \cite{adibuzzaman2016closing}. Our null hypothesis is: that there is no significant difference in outcomes for delirium patients under different drug-group in the ICU. If successful, our research work should help clear the common controversy over prescribing APDs as well as shed light on the underlying causal mechanism triggering Delirium in ICU patients. In other words, we propose the following specific aims. 
\begin{enumerate}
    \item Create and curate three cohorts for patients with Delirium in the ICU from MIMIC EHR data. 
    \item Develop structural causal models (SCM) with the domain expertise to integrate clinical knowledge and probabilistic information from the data to estimate the causal effect of interventions. 
    \item Validate the models with statistical methods and independent data sets.
\end{enumerate}

\begin{figure}[htbp]
    \centering
    \includegraphics[width=0.8\textwidth]{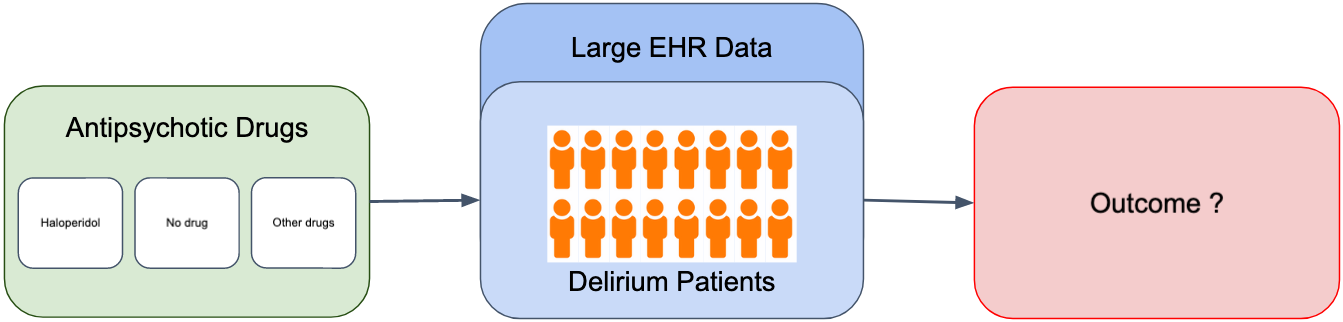}
    \caption{Target observational study from large EHR data}
    ~\label{fig:target-trial}
\end{figure}

% lit review (medical for Delirium and CS for the pipeline)
Epidemiologists have continuously involved causal inference tools, such as causal structure learning algorithms, in identifying underlying causal structures. The process is impactful since it generates a causal model based on the information available (data, literature, expertise), leading to a better understanding of the disease ecosystem; and estimates causal effects based on that. This creates a potential to explore Delirium-treatment-related controversies through observational datasets. Different studies have taken different paths; few studies \cite{vitolo2018modeling,adegunsoye2019computed} have used specific SLA algorithms to detect a causal DAG applicable for a targeted research question, whereas others \cite{schluter2019impact,arendt2016placental} have assumed the causal structure from literature, and validated them using datasets available. We plan to create a similar computational pipeline for Delirium patients in the ICU inspired by these.

%=====================================================
\section{Method}

% Dataset MIMIC
To create a data cohort on Delirium patients in the ICU, along with relevant covariates, we seek help from MIMIC-III \cite{johnson2016mimic}, a publicly available large electronic healthcare dataset. MIMIC-III is curated for twelve (12) years (2001-2012) and holds information on around 53k distinct hospital admissions with around 40k distinct patient histories. The database is well-maintained, de-identified, and open for researchers (with necessary and relevant access protocol) to explore and investigate.

% Data Mining & Analysis
The general process starts with appropriate data mining and data preparation process. We plan to extract information regarding Delirium patients (based on relevant ICD-9 code) and related covariates (decided upon exploring literature). We then move forward with the data analysis protocol, which consists of three (3) types of analysis:

\begin{itemize}
    \item Exploratory analysis: to explore data distribution and dataset properties
    \item Machine learning-driven analysis: to infer primary point of interest (i.e., primary outcome) based on all available covariates, as a standard approach to prediction
    \item Causal analysis:
    \begin{itemize}
        \item Causal structure generation: to regenerate underlying causal model through various structure learning algorithms
        \item Causal effect estimation: to evaluate the `true' causal effect of treatment on our defined points of interests
    \end{itemize}
\end{itemize}

%=====================================================
\section{Results}

This section describes our data curation protocol in detail, along with data exploration and analysis. We present our general findings based on those steps taken.

\subsection{Covariate Selection}
% Define the process goals and expectations
We start the process by defining the research questions \textit{(Is Haloperidol better at treating Delirium patients in the ICU, compared to no antipsychotics or other antipsychotics, such as Ziprasidone, Olanzapine, etc.?)}. We formulate this question based on controversies present in existing literature \textit{(described in the background section)}. Our null hypothesis is that there is no significant difference in target outcomes for Delirium patients under different antipsychotics treatment groups in the ICU. We define the treatment as the antipsychotics prescribed after being diagnosed with Delirium in the ICU, with three different arms (Haloperidol, no antipsychotics, and other antipsychotics). Our primary outcomes are (1) patient death in hospital and (2) patient death timeline (death in 30 days / 90 days / a year / survived more than a year). Our secondary outcomes are (1) length of stay in the ICU and (2) time put in mechanical ventilation. A total of fifty (50) relevant covariates are explored and marked, which are closely correlated with our points of interests (primary and secondary outcomes) for Delirium patients in the ICU. However, due to the lack of availability of all covariates in the observational dataset, we opt for the most significant twenty-eight (28) covariates, as listed in \autoref{tab:mimic-del-features}. Here, the drug group (Haloperidol, no drug, other drugs) is the treatment provided. Primary outcomes are death in hospital \& death timeline, and secondary outcomes are the length of stay \& time in mechvent.

\begin{table}
    \centering
    \begin{tabular}{|c|c|c|c|}
        \hline
        sex & age & race & icd9 codes \\
        \hline
        sofa & apsiii & surgery & pneumonia \\ 
        \hline
        sepsis & dementia & alzheimers & depression\\
        \hline
        anxiety & met. acidosis & airway obs. & copd \\
        \hline
        liver disease & heart disease & mechvent. & mechvent. count\\
        \hline
        time to mechvent. & \textbf{drug group} & drug categories count & drug timelength\\
        \hline
        \textbf{death in hospital} & \textbf{death timeline} & \textbf{length of stay} & \textbf{time in mechvent.}\\
        \hline
    \end{tabular}
    \caption{Features in MIMIC-Delirium}
    ~\label{tab:mimic-del-features}
\end{table}

These all together defined a target trial that we plan to emulate. The target trial is inspired by existing RCTs done on delirium patients to find the effects of antipsychotics and is designed to minimize the effect of confounding variables and (selection) bias. Based on these, we start our data curation process from the MIMIC-III dataset.

\subsection{Data Curation Process}
To determine eligible Delirium patients, we look into patients with ICD-9 code 293.0 (Delirium due to conditions classified elsewhere) \cite{icd9data}. We extract relevant information about the patients from admissions, icu\_stays, and diagnoses\_icd table to form the base dataset. We then infuse it with information from cptevents, d\_icd\_diagnoses and prescriptions tables, and other views presented in the public repository of the database (sofa, apsiii, ventdurations) \cite{mimic-sofa}. We merge all information together to create our target dataset of 1398 patients. We name this curated dataset as \textit{MIMIC-Delirium} for future references.
% X above: http://www.icd9data.com/2015/Volume1/290-319/290-294/293/293.0.htm

% Figure on the graph of inclusion/exclusion criteria (take an idea from Osmani bhai's and Babar Khan's paper)

\begin{figure}[htbp]
    \centering
    \includegraphics[width=\textwidth]{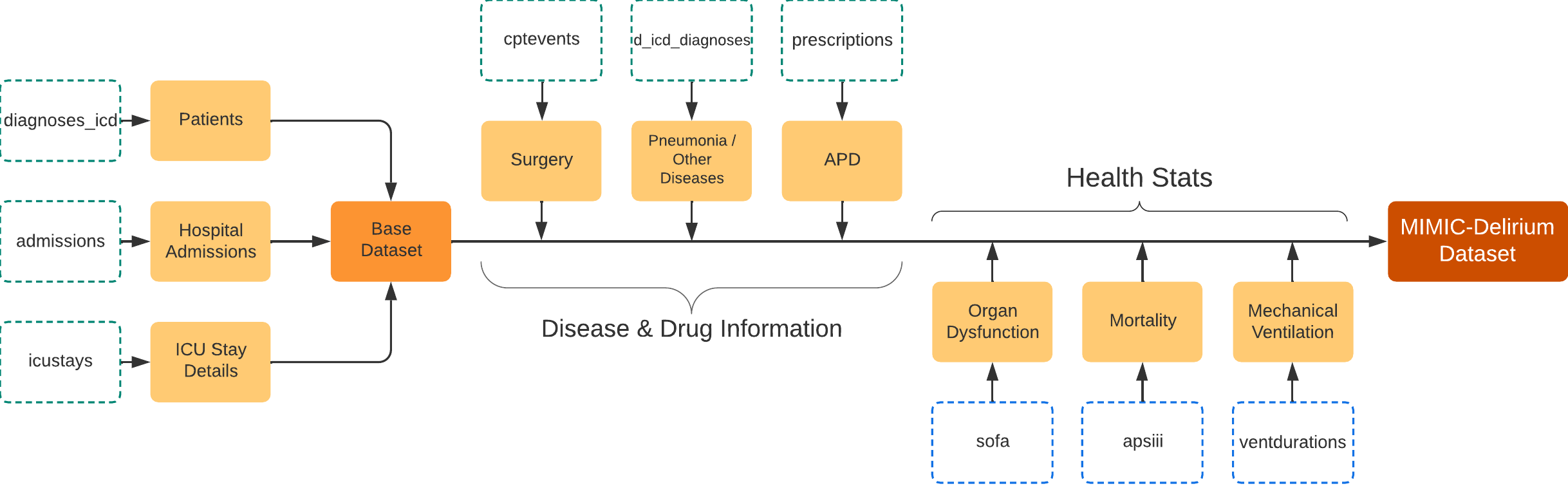}
    \caption{Data mining protocol (simplified)}
    ~\label{fig:delirium_data_mining}
\end{figure}

\subsection{Data Overview \& Exploratory Insights}

\begin{figure}[htbp]
    \centering
    \begin{minipage}{0.35\linewidth}
        \centering
        \includegraphics[width=\linewidth]{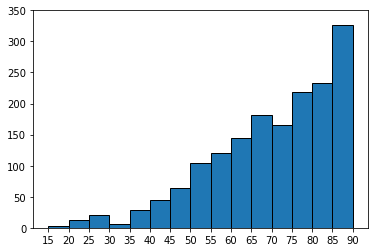}
    \end{minipage}
    \hspace{0.5cm}
    \begin{minipage}{0.35\linewidth}
        \centering
        \includegraphics[width=\linewidth]{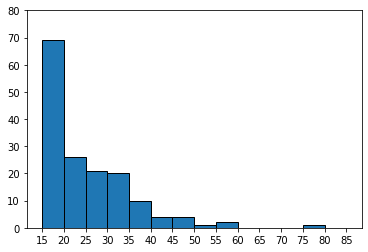}
    \end{minipage}
    \caption{Data distribution on age in years \textit{(left)} and length-of-stay in days \textit{(right)}}
    ~\label{fig:mimic-del-graph}
\end{figure}

After our data curation to create MIMIC-Delirium dataset, we successfully extract 1671 ICU stays with 1445 hospital admission counts on 1398 unique patients and their relevant 28 covariate information. In terms of treatment provided in the ICU, we found 681 \textit{(40.75\%)} were given Haloperidol, 528 \textit{(31.60\%)} were given other antipsychotics and 462 \textit{(27.65\%)} were given no antipsychotics. In terms of outcome, 311 \textit{(18.61\%)} had death in 30 days, 108 \textit{(6.46\%)} had death in 90 days, 175 \textit{(10.47\%)} had death in a year, and 253 \textit{(15.14\%)} survived at least a year (information on 821 \textit{(49.13\%)} were unknown). Among the common associated diseases in the ICU, 375 \textit{(22.44\%)} had Sepsis, 484 \textit{(28.96\%)} had Pneumonia, 1035 \textit{(61.94\%)} had (a variation of) heart diseases, and 97 \textit{(5.80\%)} had (a variation of) liver diseases. \autoref{fig:mimic-del-graph} shows the general data distribution on age in years \textit{skewed to right since Delirium is frequent in elderly population} and length-of-stay in days \textit{(skewed to left since higher number of ICU stay is severe and rare)}.

Additionally, we had findings such as
mean length-of-stay and max length-of-stay is higher for patients in the Haloperidol drug group,
most patients, who were given multiple APD, were given Haloperidol,
The Haloperidol group has a higher death rate in a year than the other two groups, etc.

For the statistical analyses, we conducted a one-way between-subjects ANOVA to compare the effect of the drug group on length of stay in Haloperidol, no drug, and other drugs group. With $p < 0.05$, we found a significant effect of the drug group on the length-of-stay. Post hoc comparisons by the Tukey HSD test indicate that the mean score for the Haloperidol group (mean: 7.47, deviation: 8.55) was significantly higher compared to no drug group (mean: 4.12, deviation: 5.66) and other drugs group (mean: 5.44, deviation: 6.14).

\subsection{Predictive Analysis on MIMIC-Delirium dataset}

\begin{figure}[htbp]
    \centering
    \includegraphics[scale=0.8]{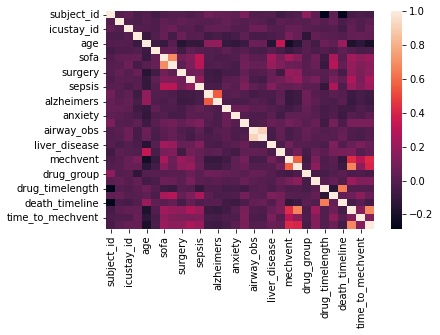}
    \caption{Correlation heatmap of MIMIC-Delirium}
    ~\label{fig:mimic-del-heatmap}
\end{figure}

Before our deep dive into causal exploration, we briefly explored the MIMIC-Delirium dataset for predictive analysis. We employed standard supervised classification algorithms on the complete dataset, with all 24 covariates (discarding the output features) as features and death in hospital as the label. We deployed 10-fold cross-validation with Logistic Regression, Support Vector Machine, and XGBoost algorithm. Mean accuracy with Logistic Regression is 89.71\%, mean accuracy with SVM is 89.11\%, and test-mlogloss-mean for XGBoost (with 50 rounds of boosts) is 0.2724. For XGBoost, we also find that length-of-stay and age have the highest impact in predicting outcome death in this case, which is self-explanatory.  \autoref{fig:mimic-del-heatmap} shows the general correlation between features as a heatmap.

\subsection{Causal Analysis on MIMIC-Delirium dataset}

Our causal analysis is built upon two steps: (1) causal structure generation and (2) causal effect estimation (based on causal structure generated).

\subsubsection{Causal Structure Generation}

\begin{figure}[htbp]
    \centering
    \includegraphics[width=\textwidth]{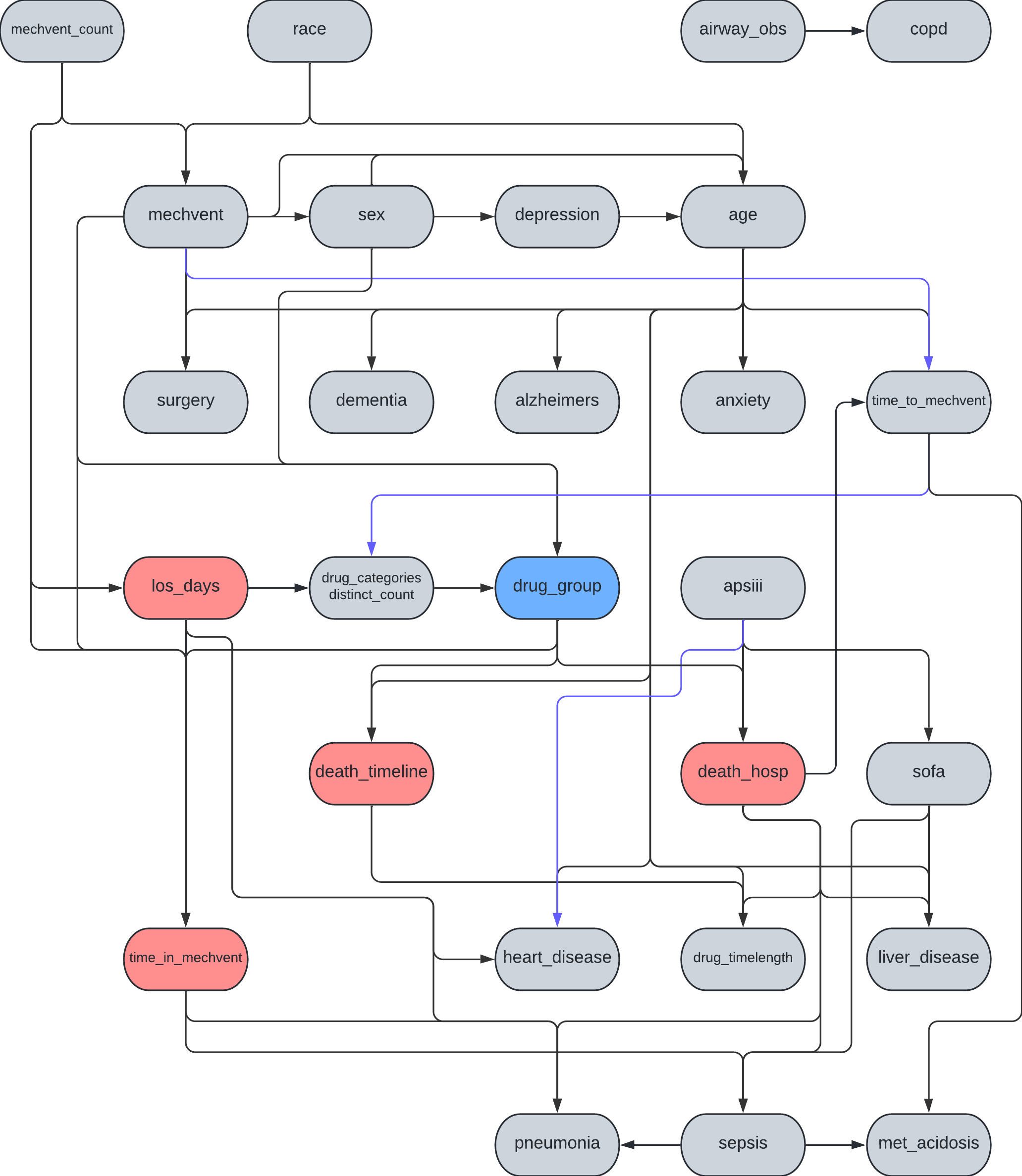}
    \caption{Combined Causal Graph for Delirium in the ICU \textit{(blue: treatment, red: primary and secondary outcomes)}}
    ~\label{fig:mimic-del-causal-dag}
\end{figure}

To generate the most feasible underlying causal structure from the MIMIC-Delirium dataset, we rely on causal structure learning algorithms (SLA), with assumptions of causal sufficiency and faithfulness. Specifically, we apply eight (8) causal structure learning algorithms: (1) PC, (2) FCI, (3) GES, (4) GIES, (5) GDS, (6) LINGAM, (7) MMHC, and (8) MMTABU, with help from existing R libraries: (1) pcalg \cite{kalisch2012causal,hauser2012characterization} and (2) pchc \cite{pchc}. With the application of these SLAs, we have eight (8) individual causal graphs. However, we apply majority voting to each edge to merge all this information together. This merging defines an edge as being present in the final graph if it is present in more than 50\% cases (more than four graphs). Although this is a straightforward and naive solution to merge multiple causal graphs, we employ this ensembling method since no standard has been established in the literature yet. \autoref{fig:mimic-del-causal-dag} shows the final merged causal graph generated.

\subsubsection{Causal Effect Estimation}

With the causal structure generated, we now focus on causal effect estimation. For this purpose, we employ the pipeline proposed by Microsoft \textbf{Do-Why} library \cite{dowhy}: 

\begin{itemize}
    \item Modeling
    \item Identification
    \item Estimation
    \item Refutation
\end{itemize}

With modeling completed as part of the causal structure generation step, we now focus on causal effect identification and estimation. Based on the causal structure generated, we identify the conditional probability equation for the four target outcomes. Specifically, we express the do-calculus operations \cite{4_bareinboim2016causal} in order to `virtually' manipulate the outcomes. The do-calculus equations are presented below:

{\footnotesize
\begin{itemize}
    \item $P(death\_in\_hosp | do(drug\_group)) = \sum_{age} P(death\_in\_hosp | drug\_group, age) P(age)$
    \item $P(death\_timeline | do(drug\_group)) = \sum_{age} P(death\_timeline | drug\_group, age) P(age)$
    \item $P(los\_days | do(drug\_group)) = \sum_{heart\_disease, mechvent} P(los\_days | drug\_group, heart\_disease, mechvent) P(heart\_disease, mechvent)$
    \item $P(time\_in\_mechvent | do(drug\_group)) = \sum_{age, mechvent} P(death\_in\_hosp | drug\_group, age, mechvent) P(age, mechvent)$
\end{itemize}}

We now find the causal effect estimates based on these causal expressions identified. In \autoref{tab:mimic-del-estimations}, we present the causal effect estimations, as Average Treatment Effects (ATE), for treatment, aka, drug group on the four target outcomes. As shown in the table, the causal effect of treatment on death in Delirium and death timeline is very close. However, any drug, Haloperidol (1.8372) and other drugs (1.6102), does much better in reducing hospital length of stay compared to the no drug patient group (-0.0533). In addition to that, any drug performs better (8.1912) in reducing time in mechanical ventilation compared to no drug (4.4827), and Haloperidol does better (12.3007) than any other drugs (8.1912).

\begin{table}
    \begin{tabular}{|l|llll|}
    \hline
    & \multicolumn{4}{c|}{\textbf{Causal effect of drug group on:}} \\ \hline
    & \multicolumn{1}{l|}{\textbf{death in}} & \multicolumn{1}{l|}{\textbf{death}} & \multicolumn{1}{l|}{\textbf{length of stay}} & \textbf{time in} \\
    & \multicolumn{1}{l|}{\textbf{hospital}} & \multicolumn{1}{l|}{\textbf{timeline}} & \multicolumn{1}{l|}{\textbf{in days}} & \textbf{mech. vent.} \\ \hline
    \textbf{Hal. vs. No Drug} & \multicolumn{1}{l|}{0.0310} & \multicolumn{1}{l|}{-0.1291} & \multicolumn{1}{l|}{1.8372} & 12.3007 \\ \hline
    \textbf{Other Drug vs. No Drug} & \multicolumn{1}{l|}{0.0216} & \multicolumn{1}{l|}{0.0373} & \multicolumn{1}{l|}{1.6102} & 8.1912 \\ \hline
    \textbf{Hal. vs. Other Drug} & \multicolumn{1}{l|}{0.0113} & \multicolumn{1}{l|}{-0.1386} & \multicolumn{1}{l|}{-0.0533} & 4.4827 \\ \hline
    \end{tabular}
    \caption{Outcomes estimation in Average Treatment Effects (ATE)}
    ~\label{tab:mimic-del-estimations}
\end{table}

We now move to the final stage of causal effect estimation, which is the refutation of the estimated effect. We do so in four different steps: by adding a random common cause to the causal model, adding an unobserved common cause to the causal model, using a Placebo treatment, and using a subset of data. The expectation for these four is that:
\begin{itemize}
    \item \textbf{Adding a random common cause}: should not change the estimated outcome from before since this should be adjusted by use of do-calculus expressions
    \item \textbf{Adding an unobserved common cause}: should change the estimated outcome from before since the unobserved confounder induces non-removable biases in the system
    \item \textbf{Using a placebo treatment}: should be close to zero since placebo treatment should not have any impact on the outcome
    \item \textbf{Using a subset of data}: should not change the estimated outcome from before since underlying data distribution did not change
\end{itemize}

Application of these four steps results in the following values, which also align with our expectations for a stable causal model and estimated effect:

\begin{itemize}
    \item Estimated effect: 0.0309
    \item Add a random common cause: 0.0310
    \item Add an Unobserved Common Cause: 0.0262
    \item Use a Placebo Treatment: 0.0003
    \item Use a subset of data: 0.0319
\end{itemize}

\section{Discussion}

% General discussion, study findings, strengths, weaknesses, future works.
We have explored a potential observational study on Delirium patients in the ICU in this study. Our curated dataset is analyzed through two lenses: regular observational analysis and `simulated' randomized controlled trial through the structural theory of causation. We have multiple novel contributions to this research work:

\begin{itemize}
    \item Our observational study creates a prospective data cohort \textit{(MIMIC-Delirium)} for Delirium patients
    \item Data properties for \textit{MIMIC-Delirium} provides insight into the general patient demography in the ICU
    \item Machine learning-driven analysis on \textit{MIMIC-Delirium} presents usage of prediction-based computational modeling and these algorithm's general strengths
    \item Causal analysis on \textit{MIMIC-Delirium} found:
    \begin{itemize}
        \item No significant impact (X) of Antipsychotics choice in one of the primary outcomes, death in hospital
        \item No significant impact (X) of Antipsychotics choice in length of stay in the ICU; however, usage of any drug shows better outcome (X) compared to that with no drugs
        \item Haloperidol performs better (X) in affecting time in mechanical ventilation, compared to the similar impact of usage of other drugs or no drugs,
    \end{itemize}
\end{itemize}

Our study relies on a few underlying assumptions. We assume that the Delirium patients in the ICU represent general Delirium demography since it occurs more frequently (~80\% cases in ICU) in the ICU compared to other traditional medical settings. Additionally, in generating the causal structure, we did not incorporate any background knowledge from peer-reviewed literature because of the existing controversies over the usage and benefits of Antipsychotics on the Delirium population \textit{(discussed in the background section)}. One of the critical limitations of our study is the lack of involvement of experienced physicians actively working in the ICU. Their involvement can aid in disputing general confusion in different parts of the study; however, bias from their understanding needs to be handled by involving multiple physicians. This limitation can be mitigated in future work. 

In summary, our proposed analysis and pipeline create pathways for similar studies, especially in the healthcare research domain. The abundance of curated large electronic healthcare data presents a potential to find unexplored insights in a specific population group. Causal inference, especially the structural theory of causation, holds the potential to handle such research questions, look for causal insights, and report them appropriately.

% CURRENT VERSION
% \input{sections/latest_main_file}

\bibliography{sample}

% \appendix
% \section*{Appendix A.}

% Some more details about those methods, so we can actually reproduce
% them.  After the blind review period, you could link to a repository
% for the code also.  

\end{document}